\documentclass[10pt,twocolumn,letterpaper]{article}

\usepackage{wacv}              

\newcommand{\auroc}{\mbox{AU-ROC}}

\newcommand{\mad}{\mbox{MVTec\hspace{1.5pt}AD}}
\newcommand{\madtwo}{\mbox{MVTec\hspace{1.5pt}AD\hspace{1.5pt}2}}

\newcommand{\testprivate}{\mbox{\textit{TEST\textsubscript{priv}}}}
\newcommand{\testprivatemixed}{\mbox{\textit{TEST\textsubscript{priv\_mix}}}}
\newcommand{\segfscore}{\mbox{$SegF_1$}}
\newcommand{\segfscoreprivate}{\mbox{$SegF_{1,\text{priv}}$}}
\newcommand{\segfscoreprivatemixed}{\mbox{$SegF_{1,\text{priv\_mix}}$}}
\newcommand{\segfscoreaverage}{\mbox{$SegF_{1,\text{avg}}$}}

\newcommand{\vandchallenge}{\mbox{VAND\hspace{1.5pt}4.0~Challenge}}

\newcommand{\apany}{\mbox{$AP_{any}$}}
\newcommand{\apmajor}{\mbox{$AP_{major}$}}
\newcommand{\kaputtrare}{Kaputt-Rare}

\usepackage[framemethod=tikz]{mdframed}
\newmdenv[
  topline=false, bottomline=false, rightline=false,
  linewidth=2pt, linecolor=orange,
  innerleftmargin=4pt, innerrightmargin=0pt,
  innertopmargin=0pt, innerbottommargin=0pt,
  skipabove=0pt, skipbelow=0pt,
  leftmargin=-6pt
]{amznbar}

\usepackage{booktabs}
\usepackage{fontawesome5}
\usepackage{pifont}

\newcommand{\kaputttwo}{\mbox{Kaputt\hspace{1.5pt}2}}

\definecolor{wacvblue}{rgb}{0.21,0.49,0.74}
\usepackage[pagebackref,breaklinks,colorlinks,allcolors=wacvblue]{hyperref}

\def\wacvPaperID{146} 
\def\confName{WACV}
\def\confYear{2027}

\title{Detecting Defects that Matter: An Application-Driven Benchmark for Anomaly Detection in Manufacturing and Retail Logistics (\vandchallenge)}

\author{
Lars Heckler-Kram\textsuperscript{1,2},
Dorian Henning\textsuperscript{3},
Ashwin Vaidya\textsuperscript{4}, 
Jan-Hendrik Neudeck\textsuperscript{1},
\\
Ulla Scheler\textsuperscript{1},
Anton Milan\textsuperscript{3},
Samet Akcay\textsuperscript{4},
Paula Ramos\textsuperscript{5},
Sebastian Höfer\textsuperscript{3}
\vspace{0.25cm} \\
{
\textsuperscript{1}MVTec Software GmbH
\textsuperscript{2}Technical University of Munich
\textsuperscript{3}Amazon
\textsuperscript{4}Intel
\textsuperscript{5}NVIDIA
}
}

\begin{document}
\maketitle

\begin{abstract}

Existing Anomaly Detection benchmarks are saturated and often unrealistic. 
As part of the \vandchallenge, we introduce a \mbox{hidden-test}, application-driven benchmark across two deployment-critical domains: industrial manufacturing and retail logistics.

In the \textit{Industrial Track} (\madtwo), the results reveal that unsupervised anomaly segmentation remains challenging: the best regular-setting method achieves only $\sim$57\% pixel-level $SegF_1$, indicating substantial room for improvement.
Zero-shot approaches trail by $\sim$15 $SegF_1$ points, confirming that task-specific training on normal data remains essential for precise defect localization.
Robustness to distribution shifts remains a key open challenge and DINOv3-backbones clearly dominate this track.

In the \textit{Retail Track} (\kaputttwo), the results reveal that (1) supervised defect detection is approaching saturation for common defect types; (2) the best off-the-shelf VLM approach trails specialized models by $\sim$28 AP, confirming that currently VLMs cannot replace fine-tuned detectors, (3) reference images did not prove helpful for top-performing approaches.
Performance collapses on rare defects (spillage $\sim$53 AP, missing units $\sim$27 AP), where the supervised ceiling is bounded by data availability.
To drive future progress in this domain, we provide a new low-prevalence retail AD dataset (\kaputtrare).

Across both tracks, computational efficiency is assessed as a first-class metric combining performance, throughput, memory, and power consumption.
We introduce a novel metric for measuring efficiency and reveal that that top-performing methods rely on heavy architectures while efficiency is largely neglected. 
Overall, we conclude that the community needs (a) more efficiency-aware method development, and (b) true anomaly detection approaches for rare defects and shifting conditions.

\noindent\href{https://sites.google.com/view/vand4-cvpr2026/challenge}{https://sites.google.com/view/vand4-cvpr2026/challenge}

\end{abstract}

\section{Introduction}
\label{sec:intro}
Visual Anomaly Detection (AD) is a computer vision task with numerous applications, ranging from medical diagnosis \cite{seebock2019_uncertainty_for_ad}, autonomous driving \cite{hendrycks2019_benchmark_anomaly_segmentation_autodriving}, video surveillance \cite{li2013_ucsd_video_ad_dataset} to industrial manufacturing \cite{bergmann2019_mvtec_ad_cvpr} and retail logistics \cite{Hoefer_2025_kaputt_dataset}.
What aligns these tasks is the sparsity of anomalous events due to the long-tail data distribution common in these settings: by definition, anomalies are rare.

For this reason, AD methods often learn to model the distribution of normal data during training and recognize anomalies as deviations from this distribution during inference.
Several approaches have been developed based on this paradigm \cite{patchcore, batzner2023_efficientad}, and Vision-Language-Models (VLMs) in particular present an exciting recent opportunity to solve visual AD requiring only limited or no normal data from the target domain at all \cite{gu2024_anomalygpt}.
%
Critically, though, the advancement of AD not only depends on method development but also on the availability of suitable benchmark datasets.
Especially in the industrial domain, datasets such as \mad\ \cite{bergmann2019_mvtec_ad_cvpr} or VisA \cite{Zou_2022_VisA} enabled the research community to develop groundbreaking new methods, but have effectively  saturated \cite{Hoefer_2025_kaputt_dataset, Heckler-Kram_2026_mvtec_ad_2_dataset} achieving up to 99.9\% \auroc\ scores.
This saturation hinders progress: genuine methodological novelty is barely distinguishable from incremental improvements, unresolved challenges such as robustness to uncontrolled settings (lighting, pose, etc.) remain unresolved, unlimited access to test data contradicts the assumption that defects may not be known.
Moreover, different benchmarks based on existing datasets vary in how they implement evaluation protocols, e.g., metrics and the amount of normal data available for training.
This lack of uniformity makes it difficult for practitioners to compare approaches and decide what approach to choose for their application at hand.
Consequently, the gap between academic research and real-world applications widens, driving the need for new benchmarks and robust evaluation protocols \cite{baitieva_2025_beyond_academic_ad_benchmarks}.

To that end, we introduce an application-driven, unified benchmark for visual AD in manufacturing and retail logistics.
We applied the benchmark as part of the \emph{Visual Anomaly and Novelty Detection Challenge} at CVPR 2026 (\vandchallenge), where numerous academic and industrial researchers contributed state-of-the-art methods for visual AD.
These methods were evaluated according to objective criteria on held-out test datasets, and we present a thorough analysis of all submissions. 
We conclude with a review of open challenges and reveal possible directions for future research.
Our contribution is three-fold:
\begin{itemize}
    \item We present a comprehensive real-world benchmark for visual AD in industrial manufacturing and retail logistics, and derive actionable conclusions about the state of the field, including evidence for performance saturation on common defects, the current limitations and (in)efficiency of VLMs, and the identification of rare defect categories as the critical unsolved problem of AD methods.
    \item We transparently elaborate on the design of the underlying challenge, enabling a clear assessment of the evaluation results under different benchmark settings.
    \item For the first time, algorithmic efficiency is considered a key criterion of an AD benchmark, which is often overlooked in academic research. We introduce a novel metric for measuring this highly deployment-relevant property. Our results highlight the need for more efficiency-focused method development.
    \item We introduce \textbf{\kaputtrare}, a novel dataset aimed at advancing methods capable of solving AD in retail logistics. 
    Featuring a defect prevalence of approximately 1\%, it constitutes a more challenging and realistic setting than Kaputt~\cite{Hoefer_2025_kaputt_dataset}, fostering the development of methods that are capable of learning (ab)normality from almost defect-free data.

\end{itemize}

\section{Related Work}
\label{sec:related_work}
\subsection{Anomaly Detection Methods}
Methods for detecting defects typically fall into one of the following paradigms: \textit{supervised}, \textit{unsupervised}, \textit{few-shot}, or \textit{zero-shot}.

Our benchmark covers the full spectrum of approaches.
Under the \textit{supervised} paradigm, methods leverage annotated defective samples; while such samples are difficult to obtain in practice due to the high cost of collection and annotation, they provide a valuable upper-bound baseline against which proper AD approaches can be measured.
AD approaches are further subdivided into \textit{unsupervised} methods, which require access only to anomaly-free reference samples, and the more challenging \textit{few-shot} and \textit{zero-shot} settings, which require minimal or no reference data from the target domain, respectively.

Especially in industrial inspection, a wide variety of unsupervised methods for AD has emerged.
These can broadly be categorized into memory bank-based \cite{patchcore, padim}, reconstruction-based \cite{Zavrtanik_DRAEM, Luo_2025_CVPR_INP-Former, Guo_2025_Dinomaly, deng2022_rd} or student-teacher-based \cite{batzner2023_efficientad, asymST} techniques.
All three families follow the same underlying principle: learning to embed normal data either explicitly (memory bank-based) or implicitly by reconstructing anomaly-free data or imitating defect-free embeddings of a known teacher network.
Frequently, synthetic defects are introduced during training to further constrain the learned feature space to anomaly-free data \cite{Zavrtanik_2022_DSR, tien2023_rd_revisited}.
At test time, deviations from the learned feature space are flagged as anomalies.

With the advent of Vision-Language Models (VLMs) such as CLIP \cite{Radford_2021_CLIP}, new paradigms for anomaly detection have been explored extensively \cite{jeong2023_winclip}, leveraging text descriptions of both defect-free and anomalous states to match features in the shared semantic space of language and vision.
Owing to the data-rich pre-training of such models, no reference data from the target domain may be required at all (\textit{zero-shot}).
Likewise, Multi-modal Large Language Models (MLLMs) such as Pixtral, GPT or Claude model families benefit from large-scale training regimes that encode deep semantic knowledge about everyday objects and also their potential normal and abnormal states directly within the model. 
This property makes MLLMs particularly appealing for domains with extreme data and item variability -- such as quality control in retail logistics \cite{Hoefer_2025_kaputt_dataset}.
However, their computational cost remains prohibitively high for real-time deployment.

The \vandchallenge\ evaluated approaches across both industrial manufacturing and retail logistics, enabling a  unique assessment of how well approaches generalize across these distinct domains in real-world applications.

\subsection{Anomaly Detection Datasets}
Several anomaly detection (AD) benchmarks have been presented and widely adopted by the research community.
Although datasets such as \mad\ \cite{bergmann2019_mvtec_ad_cvpr} or VisA \cite{Zou_2022_VisA} have fostered the development of versatile AD approaches, performance on these datasets has saturated.
Similarly, some recently proposed datasets are near-solved at the time of their publication~\cite{wang_real-iad_2024}, yet still contribute positively by broadening the range of test scenarios available for method development.
Other datasets extend the modality from 2D to 3D data \cite{bergmann2022_mvtec_3d_ad, bonfiglioli_eyecandies_2022}, while \cite{bergmann2021_mvtec_loco_ijcv} focuses on a deeper understanding of assembly semantics.
Beyond the decreasing comparability of different methods caused by saturation, a more fundamental issue persists: AD datasets commonly violate the true nature of AD by granting unlimited access to defect data during evaluation.
This can lead to methods being implicitly overfit to a  particular image resolution, feature extractor or training setup.
As a consequence, model generalization to truly unseen defects -- the core purpose of AD -- may be compromised.
For this reason, the \vandchallenge\ exclusively considered challenging datasets with restricted access to the test data, i.e., the \madtwo\ dataset \cite{Heckler-Kram_2026_mvtec_ad_2_dataset}, and the \kaputttwo\ dataset, an extension of the recently proposed Kaputt dataset \cite{Hoefer_2025_kaputt_dataset}, making the challenge a meaningful benchmark that upholds the true spirit of AD.

\subsection{Previous VAND Challenges}
The annual challenge on \textit{Visual Anomaly and Novelty Detection} has been held since CVPR 2023.
Tracking current trends in AD research, the VAND challenge continuously evolves to bridge the gap between academic research and industry.
In its first edition, it focused on zero-shot and few-shot ($\leq10$ samples) AD for industrial manufacturing.
In 2024, recent advances in VLMs were incorporated and semantic understanding of these models was benchmarked using the MVTec LOCO dataset \cite{bergmann2021_mvtec_loco_ijcv}.
Concurrently, a second track was introduced with a clear focus on robustness against changing environmental conditions such as lighting and camera viewpoints.
This track continued in 2025 \cite{heckler-kram_2025_vand3.0} but instead of to simulation, incorporated real-world imagery \cite{Heckler-Kram_2026_mvtec_ad_2_dataset}.
For CVPR 2026, AD in retail logistics was added in response to the emergence of this highly interesting yet largely unexplored domain \cite{Hoefer_2025_kaputt_dataset}.
Furthermore, for the first time, model efficiency was explicitly recognized as a criterion for evaluating contributions.
Across all editions, the challenge has provided actionable insights for both practitioners and researchers regarding the real-world applicability of AD algorithms.

\section{\vandchallenge\ and Benchmark Design}
The \vandchallenge\ featured two tracks, the \textit{industrial} and the \textit{retail} track.
Beyond detection performance, model efficiency was considered a key criterion for evaluation.

\subsection{Tracks, Settings and Award Categories}

The \textit{industrial track} was built upon the \madtwo\ dataset \cite{Heckler-Kram_2026_mvtec_ad_2_dataset}.
Evaluation was based on segmentation performance, which requires binarization of the typically continuous model predictions (\cref{subsec:metrics}).
Additionally, the \madtwo\ dataset enables assessment of robustness to varying lighting conditions -- changes that may result from distinct machine setups or the aging of illumination sources \cite{Heckler-Kram_2026_mvtec_ad_2_dataset}.
The industrial track featured two settings that differ in data accessibility during model training.
In the \emph{regular} setting, participants were required to follow the unsupervised training paradigm, i.e., training exclusively on normal images of the object category under evaluation.
In contrast, the \emph{zero-shot} setting permitted the use of any publicly available data during training but prohibited access to the \madtwo\ dataset.

The \textit{retail track} built upon the \kaputttwo\ dataset, an extension of the recently published Kaputt dataset \cite{Hoefer_2025_kaputt_dataset}. 
Kaputt/\kaputttwo\ is a large-scale dataset for AD in retail logistics.
Consequently, this track targeted benchmarking methods suitable for handling a wide variety of distinct object types and evaluated model predictions on image-level (\cref{subsec:metrics}).
In the \emph{regular} setting, teams were unconstrained in their approaches: both AD and supervised approaches were allowed, with the goal of setting a new upper baseline for the task.
Additionally, the \emph{off-the-shelf (OTS)} setting aimed to benchmark the state-of-the-art of VLMs/MLLMs in the zero-shot/few-shot setting, i.e., without fine-tuning or other weight modification of the core VLMs/MLLMs.

For both tracks, the \vandchallenge\ aligned award categories with real-world settings to incentivize application-driven method development.
To that end, submissions were not only awarded for best performance (overall, zero-shot/OTS) but also for efficiency in terms of runtime, memory and power consumption (\cref{subsec:metrics}).

\subsection{Datasets}
Datasets used in the \vandchallenge\ not only represent difficult AD scenarios based on real-world applications but also inhibit model evaluation on unseen test data, as ground truth labels were withheld and the number of evaluations strictly limited.

\paragraph{Industrial Track: \madtwo}
\madtwo\ \cite{Heckler-Kram_2026_mvtec_ad_2_dataset} is a challenging industrial AD dataset designed to overcome saturation in existing benchmarks.
It contains more than 8,000 high‑resolution images across 8 object categories (e.g., can, fabric, vial), with image resolutions ranging from 2.3 to 5 megapixels.
The dataset is split into training, validation, and multiple test sets, where training/validation include only normal images, while test sets contain both normal and anomalous samples under varying lighting conditions.
The \testprivate\ test set contains images captured under the same lighting conditions as seen during training.
In contrast, images in \testprivatemixed\ depict the same scenes but under randomly varied lighting settings, enabling robustness evaluation under distribution shifts.
Compared to other datasets (e.g. MVTec AD \cite{bergmann2019_mvtec_ad_cvpr, bergmann2021_mvtec_ad_ijcv}, VisA \cite{Zou_2022_VisA}, RealIAD \cite{wang_real-iad_2024}), \madtwo\ is smaller in classes but more complex, featuring higher intra-class variability, tiny defects, defects at the image border, and multi-object scenes.

\paragraph{Retail Track: \kaputttwo}
The Kaputt dataset \cite{Hoefer_2025_kaputt_dataset} is a large-scale benchmark for visual defect detection in retail logistics, featuring highly diverse product categories, significant pose variation compared to other AD datasets, and fine-grained annotations (defect severity, 7 defect types, segmentation masks, and item metadata).
\kaputttwo, released in conjunction with the \vandchallenge, is a held-out evaluation set of 10,000 RGB images drawn from the same distribution: it matches the original dataset in format and structure and shows comparable defect prevalence ($\sim$29\%) and defect-type distribution, but its labels remain hidden to avoid overfitting.

\subsection{Metrics}
\label{subsec:metrics}
\paragraph{Industrial Track: Segmentation F1}
Model performance in the industrial track was evaluated based on the quality of anomaly localization and segmentation using the pixel-level $F_1$ score (\segfscore), ensuring a balanced assessment of precision and recall:
\begin{equation}
\segfscore = 2 \cdot \frac{\text{precision} \cdot \text{recall}}{\text{precision} + \text{recall}}
\end{equation}
It should be noted that \segfscore\ requires the selection of a threshold for the typically continuous anomaly maps. This aspect is often not yet addressed in the scientific community, although it is essential for deployment in real-world applications.
In line with standard evaluation protocols in anomaly detection, precision and recall were computed over the complete set of pixels in the corresponding test set rather than being averaged over individual images.

Finally, the final rank $\mathcal{R}_{\text{final}}$ of submissions to the industrial track of the \vandchallenge\ accounted for both overall performance and robustness against real-world distribution shifts.
It was calculated as the average rank of a model on \testprivate\ and \testprivatemixed\ with respect to the mean \segfscore\ across all eight object categories $c$ of \madtwo:
\begin{equation}
\mathcal{R}_{\text{final}} = \frac{1}{2} [\mathcal{R}(\segfscoreprivate) + \mathcal{R}(\segfscoreprivatemixed)]
\end{equation}
with
\begin{equation}
SegF_{1,t} = \frac{1}{8}\sum_{c = 1}^8{SegF_{1,c}}(\text{\textit{TEST}}_t)
\end{equation}
where $t$ denotes the test split (\textit{priv}, \textit{priv\_mix}).
In cases where submissions achieved the same $\mathcal{R}_{\text{final}}$, the leaderboard position was determined by the smaller absolute difference between \segfscoreprivate\ and \segfscoreprivatemixed.

\paragraph{Retail Track: Average Precision}
Model performance in the retail track assessed the quality of visual defect detection in real-world retail logistics scenarios based on Average Precision for any defect (\apany). This metric considers both minor and major defects as positive samples and evaluates how well a model ranks defective items ahead of non-defective ones across all possible decision thresholds:
\begin{equation}
\apany = \sum_n (R_n - R_{n-1}) \cdot P_n
\end{equation}

Here, $P_n$ and $R_n$ denote precision and recall at threshold $n$, with $P_n = TP/(TP + FP)$ and $R_n = TP/(TP + FN)$, where true positives ($TP$) are defective items correctly identified, false positives ($FP$) non-defective items incorrectly flagged, and false negatives ($FN$) defective items missed.

In contrast to the industrial track, which evaluates pixel-level anomaly localization and segmentation, the retail track focuses on image-level defect detection and primarily measures performance as average precision on all defect types (\apany).
Since \apany\ is computed over the ranked defect probability scores submitted for the \kaputttwo\ test set, it avoids the need to select a single fixed operating threshold during evaluation while still capturing performance over the full precision-recall curve.

Ultimately, the final ranking of submissions in the retail track was determined by the \apany\ score on the \kaputttwo\ test set:
\begin{equation}
\mathcal{R}_{\text{final}} = \mathcal{R}(\apany)
\end{equation}
where higher \apany\ values indicate better overall defect detection performance. The evaluation is based on prediction files containing one defect probability score per test image, with larger scores indicating a higher likelihood that the corresponding item is defective.

\paragraph{Computational Efficiency}

In addition to predictive performance, submissions were evaluated with respect to computational efficiency to account for practical deployment constraints.
The efficiency score combines detection quality with three resource-related measures: inference throughput, peak memory usage, and average power consumption. More specifically, the score is based on the model performance $q$, the normalized throughput $\tau$, the normalized memory usage $\mu$, and the normalized power consumption $\pi$.
While higher values of $q$ and $\tau$ are desirable, lower memory and power consumption are preferred and therefore enter the efficiency score $\mathcal{E}$ through their complements:
\begin{equation}
\label{equ:efficiency_score}
\mathcal{E} = \left(q \cdot \tau \cdot (1 - \mu) \cdot (1 - \pi)\right)^{1/4}
\end{equation}
Here, $q$ denotes the task-specific performance score ($q_\text{industrial} = \segfscoreaverage = (\segfscoreprivate\ + \segfscoreprivatemixed)/2$; $q_\text{retail} = \apany$), while the raw throughput $t$ is normalized with respect to a target throughput $t_{\max}$~(30fps) using a diminishing-returns function: $\tau = t/(t + t_{\max})$.
Peak memory consumption $m$ and average power draw~$p$ are normalized relative to predefined resource budgets \mbox{$m_{\max}$~(64GB)} and $p_{\max}$~(56W): $\mu = m/m_{\max}$, $\pi = p/p_{\max}$.

The geometric mean was chosen to encourage balanced models, since a severe weakness in any individual component substantially reduces the final score. 
This ensures that models cannot achieve a high efficiency score solely through strong predictive performance or high throughput if they require excessive memory or power.
Conversely, the throughput normalization introduces diminishing returns beyond the target operating point, preventing raw speed from dominating the evaluation.
A more detailed description of the efficiency metric, including the motivation for the normalization functions and an analysis of its sensitivity to the individual components, is provided in the appendix.

\subsection{Submission Process and Evaluation Protocol}

Both \vandchallenge\ tracks were hosted on easily accessible benchmark platforms, with the industrial track evaluated via the MVTec Benchmark Server\footnote{\href{https://www.benchmark.mvtec.com}{benchmark.mvtec.com}} and the retail track via Codabench\footnote{\href{https://codabench.org/competitions/14156}{codabench.org/competitions/14156}}.
To preserve the integrity of the hidden test sets and prevent excessive tuning on the benchmark data, both tracks enforced a limited evaluation budget, namely three submissions per seven days for the industrial track and four submissions in total for the retail track.
The evaluation protocol followed the track-specific metrics described above (\cref{subsec:metrics}).

To be considered for the final leaderboard and prize eligibility, submissions were required to contribute to the community by providing a reproducible open-source code repository, including all necessary model weights or download instructions, as well as a public technical report describing the method.
Computational efficiency was assessed after the challenge deadline under a unified hardware setting as also described in the appendix.

\section{Results and Analysis}
\subsection{Industrial Track}
Only considering complete entries providing report and code, the industrial track ultimately received 9 submissions in the regular setting and 5 submissions in the zero-shot setting.
Consequently, compared to the previous VAND challenge in 2025 \cite{heckler-kram_2025_vand3.0}, the number of teams competing in this track increased by 40\%.

\paragraph{Regular Setting}

\begin{table}[t]
\centering
\caption{\vandchallenge\ industrial track (\madtwo) leaderboard under the regular setting.}
\label{tab:results_inudstrial_regular}
\resizebox{\columnwidth}{!}{%
\begin{tabular}{c l c c c c}
\toprule
$\mathcal{R}_{\text{final}}$ & Method & \segfscoreprivate & \segfscoreprivatemixed & Report & Code \\
\midrule
1  & SuperADD        & 57.42 & 54.35 & \href{https://arxiv.org/abs/2605.14808}{\ding{72}}& \href{https://github.com/LukasRoom/SuperADD}{\faGithub}\\
2  & ISVL++          & 58.31 & 53.75 & \href{https://github.com/ISVL119/isvl-plusplus/blob/main/isvl++.pdf}{\ding{72}}& \href{https://github.com/ISVL119/isvl-plusplus}{\faGithub}\\
3  & Forgis\_baseline& 49.02 & 45.80 & \href{https://github.com/Forgis-Labs/for-vand-2026/blob/main/paper.pdf}{\ding{72}}& \href{https://github.com/Forgis-Labs/for-vand-2026}{\faGithub}\\
4  & MSAD& 48.54 & 45.50 & \href{https://github.com/yyyangcoder/MSAD/blob/main/paper.pdf}{\ding{72}}& \href{https://github.com/yyyangcoder/MSAD}{\faGithub}\\
5  & DL-SubspaceAD   & 46.88 & 45.36 & \href{https://github.com/riips/VAND4.0_IndustrialTrack_Regular/blob/main/report.pdf}{\ding{72}}& \href{https://github.com/riips/VAND4.0_IndustrialTrack_Regular}{\faGithub}\\
6  & vandseg\_4       & 46.48 & 40.61 & \href{https://github.com/Jessejx/VAND2026-Industrial-Anomaly-Detection-ADAD/blob/main/VADN2026.pdf}{\ding{72}}& \href{https://github.com/Jessejx/VAND2026-Industrial-Anomaly-Detection-ADAD/tree/main}{\faGithub}\\
7  & CAT             & 44.26 & 39.49 & \href{https://github.com/VisualScienceLab-KHU/VAND2026_CAT/blob/main/vand4.0.pdf}{\ding{72}}& \href{https://github.com/VisualScienceLab-KHU/VAND2026_CAT}{\faGithub}\\
8  & ST              & 41.53 & 33.97 & \href{https://gitlab.com/drti/vand4_safrantech/-/blob/main/VAND4_Regular_setting_ST.pdf}{\ding{72}}& \href{https://gitlab.com/drti/vand4_safrantech}{\faGithub}\\
9  & GBE\_AD         & 36.36 & 31.60 & \href{https://drive.google.com/file/d/19XTLOWqA9sgkufbplIEHNMdUOCmQQ75N}{\ding{72}}& \href{https://github.com/YonatanZehavi319/vand-2026}{\faGithub}\\
\midrule
VAND\hspace{1.5pt}3.0 & ISVL \cite{Wang_2025_ISVL} & 53.81 & 51.43 & \href{https://github.com/ISVL119/isvl/blob/main/vand3.0.pdf}{\ding{72}} & \href{https://www.github.com/ISVL119/isvl}{\faGithub} \\
\bottomrule
\end{tabular}
}
\end{table}
Though rules for model design in the regular setting (\cref{tab:results_inudstrial_regular}) were stricter than in the previous VAND edition (VAND\hspace{1.5pt}3.0 \cite{heckler-kram_2025_vand3.0}), i.e., prohibiting category-specific model selection, last year's top-performing submission ISVL \cite{Wang_2025_ISVL} was outperformed by more than 2.5\%-points and nearly 3\%-points in terms of \segfscoreprivate\ and \segfscoreprivatemixed, respectively (\cref{tab:results_inudstrial_regular}).
Apart from improvement in pre- and postprocessing, the winning solution \textit{SuperADD} \cite{Roming_2026_SuperADD} adapts last year's submission \textit{SuperAD} \cite{Zhang_2025_SuperAD} by incorporating a \mbox{DINOv3} \cite{simeoni2025dinov3} backbone and advancing the subsampling process when building a memory bank of normal features.
Similarly, second place \textit{ISVL++} \cite{Wang_2026_ISVL++} adopts \textit{ISVL} \cite{Wang_2025_ISVL} with a DINOv3 backbone \cite{simeoni2025dinov3} and utilizes synthetic anomaly data for training a segmentation head on top of INP-Former \cite{Luo_2025_CVPR_INP-Former} and for threshold estimation.

\paragraph{Zero-Shot Setting}

\begin{table}[ht]
\centering
\caption{\vandchallenge\ industrial track leaderboard (\madtwo) under the zero-shot setting.}
\label{tab:results_industrial_zeroshot}
\resizebox{\columnwidth}{!}{%
\begin{tabular}{c l c c c c}
\toprule
$\mathcal{R}_{\text{final}}$ & Method & \segfscoreprivate & \segfscoreprivatemixed & Report & Code \\
\midrule
1 & BALM        & 42.64 & 41.45 & \href{https://github.com/Serena-cooper/BALM/blob/main/BALM.pdf}{\ding{72}}& \href{https://github.com/Serena-cooper/BALM}{\faGithub}\\
2 & ProtoFuse   & 30.50 & 30.44 & \href{https://github.com/riips/VAND4.0_Industrial_Zeroshot_ProtoFuse/blob/main/technical_report.pdf}{\ding{72}}& \href{https://github.com/riips/VAND4.0_Industrial_Zeroshot_ProtoFuse}{\faGithub}\\
3 & SPR (lin)         & 23.96 & 24.05 & \href{https://github.com/wwwlin0213/SPR/blob/main/SPR.pdf}{\ding{72}}& \href{https://github.com/wwwlin0213/SPR}{\faGithub}\\
4 & sim2        & 24.55 & 22.84 & \href{https://github.com/yongheng0011/ASCLIP/blob/main/VAND4_0.pdf}{\ding{72}}& \href{https://github.com/yongheng0011/ASCLIP}{\faGithub}\\
5 & EthMuSAM    & 23.57 & 22.15 & \href{https://github.com/jubaer36/EthMuSAM/blob/main/EthMuSAM.pdf}{\ding{72}}& \href{https://github.com/jubaer36/EthMuSAM}{\faGithub}\\
\bottomrule
\end{tabular}%
}
\end{table}

In the zero-shot setting (\cref{tab:results_industrial_zeroshot}), \textit{BALM} \cite{Gao_2026_BALM} extends AnomalyVFM \cite{Fucka_2026_AnomalyVFM} by integrating the task of threshold estimation more intensively into model training.
SAM \cite{kirillov_2023_segany} is used for refining resulting binary segmentation maps.
Similar to the regular setting, recent methods are built upon by updating the feature extractor.
For example, among other modifications, ProtoFuse \cite{Nanaumi_2026_ProtoFuse} or SPR \cite{Tan_2026_SPR} utilize \textit{VisualAD} \cite{Hou_2026_VisualAD} with a DINOv3 instead of a \mbox{DINOv2} backbone.

\paragraph{Analysis}
Results of the industrial track share clear patterns reflecting current trends in anomaly detection research for visual inspection in manufacturing.

\textit{DINOv3 takes the lead.}
The dominant trend, especially observable in the regular setting, is the strong reliance on DINOv3 \cite{simeoni2025dinov3} features.
This suggests a shift from designing increasingly complex anomaly-detection architectures toward exploiting very strong general-purpose visual representations and combining them with comparatively simple anomaly-scoring mechanisms.
As an example, this is observable in \textit{DL-Subspace AD} \cite{Ito_2026_DLSubspaceAD}, which inserts a ViT 7B/16-based DINOv3 \cite{simeoni2025dinov3} backbone into the original method \textit{Subspace AD} \cite{Lendering_2026_SubspaceAD}, enabling using compression and reconstruction by principle component analysis for anomaly detection.

\textit{Resolution matters.}
As in VAND\hspace{1.5pt}3.0, tiling and multi-scale processing is utilized to handle the large image sizes contained in \madtwo, which reflects the difficulty of detecting subtle defects in high-resolution images.
Certainly, processing images as tiles or at multiple or high resolutions brings a computational burden, which might significantly harm model efficiency (\cref{subsec:efficiency}).

\textit{Binarization by Morphology and Deep Learning.}
Many submissions use morphological operations and connected-component filtering in addition to segmentation refinement by SAM \cite{kirillov_2023_segany} modules to obtain binary anomaly segmentation maps.
On the one hand, this shows how classic and deep learning-based approaches can be leveraged together.
On the other hand, the question arises whether threshold selection/binarization can be integrated even further into model design without the need for empiric post-processing strategies - regardless of their manner.

\textit{Robustness diminishes by target-domain access.}
Robustness to illumination shifts remains a central difficulty.
Though some methods such as \textit{DL-Subspace~AD} \cite{Ito_2026_DLSubspaceAD} are comparably robust against changing lighting conditions, most approaches in the regular setting show a significant performance difference between \segfscoreprivate\ and \segfscoreprivatemixed.
Conversely, the performance gap between the test sets with seen (\testprivate) and partly unseen (\testprivatemixed) lighting conditions in the zero-shot setting is usually smaller than 1.5\%-points.
Hence, the question arises whether increased robustness can be induced into methods that have access to data from the target domain, i.e., normal samples from the object category to be investigated.
A deeper understanding of the underlying mechanisms leading to robustness against or sensitivity to different lighting is still needed, potentially revealing measures applicable to other kind of distribution shifts induced by changes in image acquisition.

\subsection{Retail Track}
The retail track attracted 74 unique participants who submitted 286 predictions on the \kaputttwo\ test set.
Out of the 286 submitted predictions, 17 were submitted as OTS predictions.

\paragraph{Regular Setting}
\begin{table}[ht]
\centering
\caption{\vandchallenge\ retail track (\kaputttwo) leaderboard under the regular setting.}
\label{tab:results_retail_regular}
\begin{tabular}{c l c c c}
\toprule
$\mathcal{R}_{\text{final}}$ & Team & \apany & Report & Code \\
\midrule
1  & dinovision      & 90.84 & \href{https://github.com/dorianhenning/VAND2026_Kaputt/blob/add-technical-report/technical_report.pdf}{\ding{72}} & \href{https://github.com/teamdinovision/VAND2026_Kaputt}{\faGithub} \\
2  & sandeepsaluru   & 89.30 & \href{https://github.com/durga-sandeep/kaputt-defect-detection-cvpr2026/blob/main/techical_report.pdf}{\ding{72}} & \href{https://github.com/durga-sandeep/kaputt-defect-detection-cvpr2026}{\faGithub} \\
3  & xie233          & 84.32 & \href{https://github.com/dorianhenning/CVPR2026_Retail_Track_Solution/blob/add-technical-report/technical_report.pdf}{\ding{72}} & \href{https://github.com/HowYear/CVPR2026_Retail_Track_Solution}{\faGithub} \\
\bottomrule
\end{tabular}
\end{table}

\Cref{tab:results_retail_regular} shows submissions to the retail track under the regular setting that included a report and code for reproducibility.
The best submission achieved an \apany\ of 90.84.
However, it is worth mentioning that the entire top-30 spanned less than 1.6~AP points (89.27--90.84) \footnote{\href{https://www.codabench.org/competitions/14156/\#/results-tab}{codabench.org/competitions/14156/\#/results-tab}}.
The winning team \textit{dinovision}~\cite{Tao_2026_dinovision_regular} uses an ensemble of three large vision transformer backbones (\mbox{DINOv2-Large}~\cite{oquab2023dinov2}, \mbox{DINOv3-Large}, \mbox{DINOv3-Huge}~\cite{simeoni2025dinov3}) combined with an increase of test-time augmentation of 56-views, representing approximately 3--5$\times$ the inference cost of a single-model approach.
The runner-up \textit{sandeepsaluru}~\cite{Saluru_2026_sandeepsaluru_regular} used a two-member \mbox{DINOv3-Large}~\cite{simeoni2025dinov3} ensemble at approximately 2$\times$ the cost of a single-model.
Both approaches relied on supervised training on Kaputt~\cite{Hoefer_2025_kaputt_dataset} data and leveraged patch-level representations of DINOv3~\cite{simeoni2025dinov3} for defect classification.
In particular, reference images were rarely used by any of the methods, suggesting their limited usefulness for top-performing approaches.

\paragraph{Off-the-Shelf VLM Setting}

  \begin{table}[ht]
  \centering
  \caption{\vandchallenge\ retail track (\kaputttwo) leaderboard under the off-the-shelf (OTS) setting.}
  \label{tab:results_retail_ots}
  \begin{tabular}{c l c c c}
  \toprule
  $\mathcal{R}_{\text{final}}$ & Team & \apany & Report & Code \\
  \midrule
  1  & rahulvippala    & 61.88 & \href{https://github.com/rvi876/vand/blob/main/tech_report.pdf}{\ding{72}} & \href{https://github.com/rvi876/vand}{\faGithub}
  \\
  2  & dinovision      & 61.83 & \href{https://github.com/dorianhenning/VAND2026_Kaputt_OTS/blob/add-technical-report/technical_report.pdf}{\ding{72}} &
  \href{https://github.com/teamdinovision/VAND2026_Kaputt_OTS}{\faGithub} \\
  3  & ant-hy          & 54.87 & \href{https://github.com/dorianhenning/PD-VLM/blob/add-technical-report/technical_report.pdf}{\ding{72}} &
  \href{https://github.com/Greywan/PD-VLM}{\faGithub} \\
  4  & amirparsa-sal   & 50.49 & \href{https://github.com/Amirparsa-Sal/Polimaly-Anomaly-Detection/blob/master/technical_report.pdf}{\ding{72}} &
  \href{https://github.com/Amirparsa-Sal/Polimaly-Anomaly-Detection}{\faGithub} \\
  5  & sandeepsaluru   & 46.75 & \href{https://github.com/durga-sandeep/kaputt-defect-detection-cvpr2026/blob/main/techical_report.pdf}{\ding{72}} &
  \href{https://github.com/durga-sandeep/kaputt-defect-detection-cvpr2026}{\faGithub} \\
  \bottomrule
  \end{tabular}
  \end{table}

The best OTS VLM submission (\cref{tab:results_retail_ots}) achieved an \apany\ of 61.88, trailing the best supervised model by approximately 29~AP points~\cite{Vippala_2026_rahulvippala_ots}.
This confirms that general-purpose VLMs cannot currently replace fine-tuned classifiers for retail defect detection.
The VLMs used by the participants included Gemini~3.1~Pro~Gemini~\cite{gemini_2025}, Gemma4-31B~\cite{gemma3_2025}, Qwen3.5-27B, Qwen3.5-9B~\cite{qwen3_5_omni_2026}, InternVL~\cite{internvl2_5_2024}, and GPT-4o~\cite{gpt4o_2024}.
The key innovations that drove performance gains over a naive prompting baseline (+19.6~AP) were multi-prompt ensembles~\cite{Wan_2026_anthy_ots}, logprob-based confidence scoring (extracting continuous defect probabilities from token log-probabilities rather than parsing binary text outputs)~\cite{Tao_2026_dinovision_ots}, and structured, per-category output schemas~\cite{Vippala_2026_rahulvippala_ots}.
The top-performing approaches found that multi-prompt diversity, i.e., probing with deliberately differently phrased questions, was more effective than multi-scale or multi-model diversity~\cite{Vippala_2026_rahulvippala_ots,Wan_2026_anthy_ots,Saluru_2026_sandeepsaluru_regular}.

\paragraph{Analysis}

The Retail Track results reveal several findings relevant to visual defect detection in logistics.

\textit{Performance saturation for common defects.}
The compressed leaderboard (top-30 within 1.6~\apany) indicates that common and visually salient defect types such as deformation and penetration, are largely solved when sufficient labeled data are available.
Marginal gains at this level require computationally expensive ensembles rather than algorithmic innovation, suggesting diminishing returns from further supervised method development in these categories.

\begin{figure}[ht]
\centering
\includegraphics[width=\columnwidth]{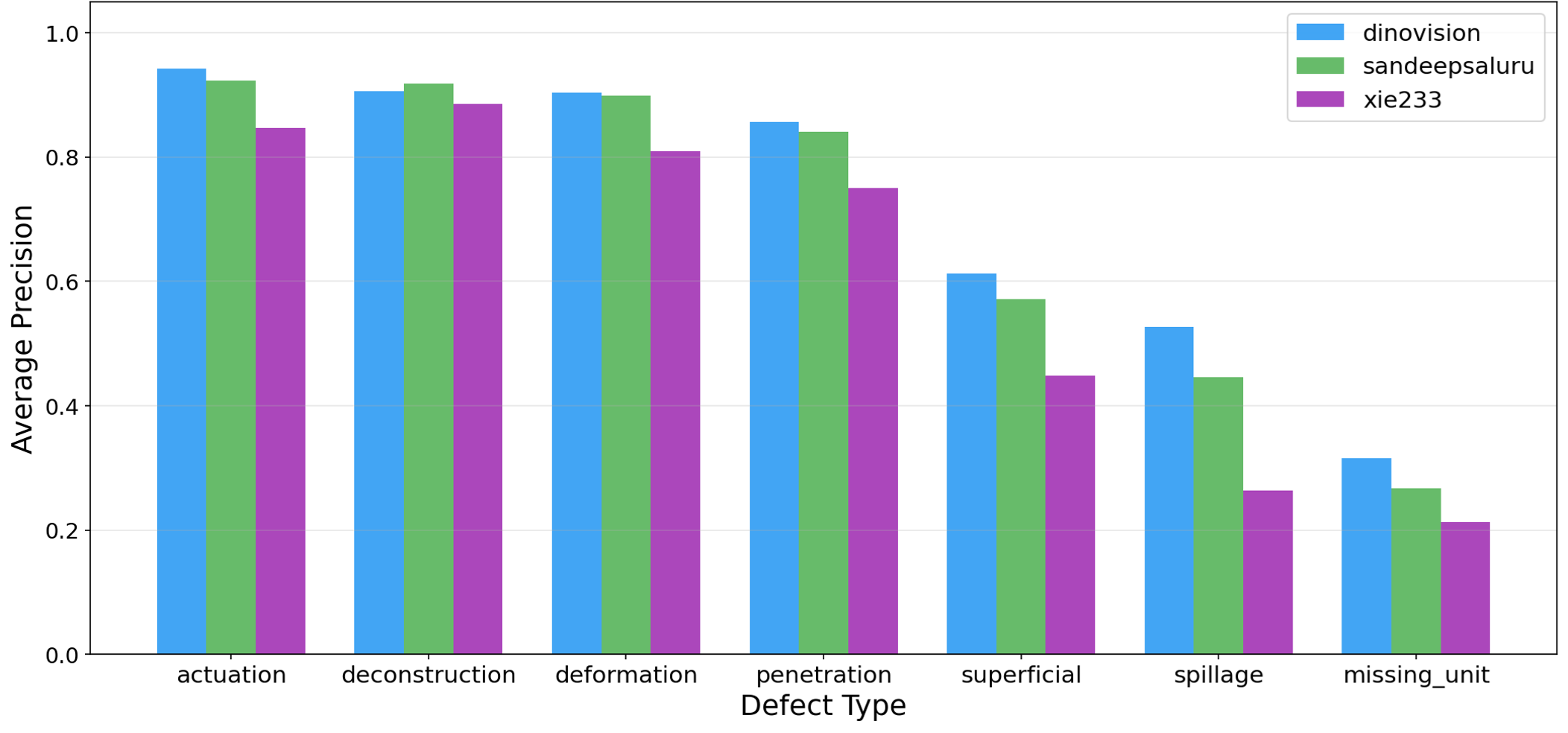}
\caption{Per-defect-type \apany\ of the top three submissions on \kaputttwo. Common structural defects are largely solved, while rare categories (spillage, missing unit) remain challenging.}
\label{fig:defect_type_ap}
\end{figure}

\textit{Failure on rare defect types.}
The per-defect-type analysis (\cref{fig:defect_type_ap}) reveals that all submissions struggle with underrepresented defect categories: spillage achieved only $\sim$53~\apany\ and missing unit $\sim$27~\apany.
These failure modes are due to the data scarcity in the training set (e.g., only 705 instances with spillages in Kaputt~\cite{Hoefer_2025_kaputt_dataset}) and from the fundamentally different visual characteristics of these defects compared to structural damage.
Notably, reference images that provide a non-defective view of the same item did not help close this gap: top-performing methods in both settings did not leverage them as a primary signal, despite the \kaputttwo\ test set providing up to three per query item.
This contrasts with the industrial track, where reference-based methods are standard, and challenges the assumption that the extreme product variety in retail logistics necessitates explicit normal references for anomaly detection.
Instead, the results suggest that current methods extract sufficient normality priors from large-scale supervised training, while the true bottleneck lies in the scarcity of rare defect examples rather than the absence of reference data.

\textit{VLMs are not yet competitive, but improving rapidly.}
The $\sim$29~\apany~gap between OTS and supervised models confirms that VLMs lack the fine-grained visual discrimination needed for retail defect detection.
However, prompt engineering techniques yielded substantial improvements, suggesting that VLMs may become increasingly viable as their visual grounding capabilities improve.
The finding that multi-prompt ensembles outperform multi-model ensembles indicates that repeated judgments from a single model is more effective than combining signals across different models.

\textit{Trends.}
The dominant architectural trend mirrors the industrial track: DINOv3-based feature extraction combined with task-specific classification heads.
Unlike industrial inspection, where spatial localization is critical, the retail track favored global image-level classification, leveraging aggregated patch features rather than local anomaly scoring.
Test-time augmentation with multiple views was universally adopted among top submissions, to counteract the pose variability present in the dataset.

\subsection{Efficiency}
\label{subsec:efficiency}

\cref{tab:efficiency_results} summarizes the efficiency results of the \vandchallenge\ for both tracks.
Efficiency was only measured for submissions that could complete \mbox{end-to-end} inference within hardware resource limits (industrial edge device with a Core Ultra Series 3 processor and 64 GB RAM).
For the retail track, the only eligible method reaches an efficiency score of 0.3473, indicating large room for improvements.
It distills soft targets from a DINOv2-ViTB/14-based teacher to a DINOv2-ViTS/14-based student without requiring any heavy VLM components~\cite{Xie_2026_xie233_regular}.
In the industrial track, SPR \cite{Tan_2026_SPR} reaches the highest efficiency score of 0.1937.
Notably, SPR has the worst performance score $q$ (\segfscoreaverage) of all methods considered, but excels in throughput, memory usage, and power consumption, which is likely due to using an image size of only $336 \times 336$ pixels.
Overall, these results highlight that the submissions focused primarily on achieving high performance while neglecting resource efficient inference.

\begin{table}[h]
\centering
\resizebox{\columnwidth}{!}{%
\begin{tabular}{c l c c c c c}
\hline
$\mathcal{R}_\text{eff}$ & Team/Method& $\mathcal{E} \uparrow$ & $\tau \uparrow$ & $\mu \downarrow$ & $\pi \downarrow$ & $q \uparrow$ \\
\hline
\hline
1 & xie233& \textbf{0.3473} & \textbf{0.1093} & \textbf{0.7160} & \textbf{0.4442} & \textbf{0.8432} \\
\hline
1  & SPR (lin)& \textbf{0.1937} & \underline{0.0993} & \textbf{0.8964} & \underline{0.4300} & 0.2401 \\
2  & Forgis\_baseline & \underline{0.1599} & \textbf{0.1185} & 0.9791 & 0.4441 & 0.4741 \\
3  & GBE\_AD& 0.1415 & 0.0221 & \underline{0.9088} & 0.4144 & 0.3398 \\
4  & BALM & 0.1414 & 0.0250 & 0.9376 & \textbf{0.3899} & 0.4205 \\
5  & CAT& 0.1350 & 0.0631 & 0.9772 & 0.4477 & 0.4188 \\
6  & ISVL++ & 0.1219 & 0.0083 & 0.9142 & 0.4441 & \underline{0.5603} \\
7  & MSAD & 0.1140 & 0.0355 & 0.9818 & 0.4446 & 0.4702 \\
8  & SuperADD & 0.1057 & 0.0041 & 0.9028 & 0.4435 & \textbf{0.5589} \\
\hline
\end{tabular}
}
\caption{Efficiency rank $\mathcal{R}_\text{eff}$ according to efficiency score $\mathcal{E}$  for the retail track (upper part; $q:=\textbf{\apany}$) and industrial track (lower part; $q:=\segfscoreaverage$). $\mathcal{E}$~is the geometric mean of normalized throughput $\tau$, normalized memory usage $\mu$, normalized power consumption $\pi$ and model performance $q$ (\cref{equ:efficiency_score}). Only methods runnable on the target hardware are shown. Best (second best) result per track and column is marked bold (underlined).}
\label{tab:efficiency_results}
\end{table}

\section{Discussion}
The \vandchallenge\ provides a comprehensive assessment of the current state of anomaly detection across two distinct application domains - visual inspection in manufacturing and retail.

\textit{Zero-shot and VLM-based methods remain far from practical deployment.}
The zero-shot industrial gap ($\sim$15~\segfscore\ below regular) and the retail OTS VLM gap ($\sim$29~\apany\ below supervised) confirm that general-purpose models cannot yet match task-specific finetuning.
However, both tracks demonstrate that prompt engineering and transfer learning from related domains can substantially close this gap, pointing to a promising trajectory for future development.

\textit{Rare defects and distribution shifts remain the primary open challenges.}
In the retail track, rare defect types (spillage, missing units) represent genuine failure modes that supervised approaches cannot solve without additional data.
In the industrial track, robustness to unseen lighting conditions causes consistent performance degradation.
Both scenarios reflect core properties of real-world AD: defects are inherently rare and acquisition conditions vary.

\textit{Efficiency deserves greater attention.}
The winning performance submission either relied on expensive ensembling (retail track) or utilized a heavy backbone (industrial track), both leading to high resource requirements and, thus, decreased efficiency.
However, for practical deployment, the efficiency-performance trade-off should be a primary consideration in future method development.
Besides, the question whether comparable anomaly detection performance can be achieved with more lightweight networks might open the room for new methodological approaches.

\section{Conclusion}
We presented the \vandchallenge\ as an application-driven benchmark for visual anomaly detection in industrial manufacturing and retail logistics.
Considering datasets holding the true nature of anomaly detection, i.e., not knowing which defects to expect at test time, across the two tracks, we established that: (1)~supervised defect detection in retail logistics is approaching saturation for common defect types but rare defects remain unsolved; (2)~off-the-shelf VLMs/zero-shot approaches trail specialized models but benefit substantially from prompt engineering, ensembles and transfer learning; (3)~improving the performance-efficiency-tradeoff is an open problem, especially prominent for detecting small and subtle defects in large images - a common setting in real-world industrial visual inspection.

\paragraph{Limitations}
The challenge design has several aspects to improve in future iterations.

\textit{Efficiency evaluation}. The efficiency track was evaluated post-hoc rather than as a live competition criterion, which limited participation. Integrating efficiency constraints directly into the submission process could incentivize more resource-aware method development.

\textit{Separation of evaluation protocols}. In the retail track, the final ranking did not distinguish between scenarios with and without reference images, making it difficult to isolate their contribution. 
Similarly, in the industrial track, zero-shot anomaly detection would benefit from stricter boundaries around what forms of test-time adaptation, test-set normalization, or cross-image comparison are permitted. Clearer rules would both guide participants and help the community converge on a unified definition of zero-shot anomaly detection with respect to allowed pretraining data.

\textit{Controlled feature extraction}. Although evaluation protocols are pre-defined, comparability between methodological approaches could be further improved by also pre-defining the feature extractor — potentially as a lightweight architecture. Alternatively, requiring ablation studies on the backbone architecture would help isolate each method's true contribution.

\paragraph{Future Work: \kaputtrare}
The challenge results demonstrate that supervised defect detection saturates for common defect types in retail logistics but fails on rare defect categories, precisely where true anomaly detection is needed.
To drive progress in this direction, we release \textbf{\kaputtrare}, a re-partitioned variant of the Kaputt dataset \cite{Hoefer_2025_kaputt_dataset} designed for the anomaly and novelty detection setting.
\kaputtrare\ features a defect prevalence of approximately 1\%, simulating a realistic low-prevalence scenario typical of quality-controlled logistics environments.
This formulation requires methods to learn (ab)normality from almost defect-free data.

We provide initial baseline results in~\cref{tab:kaputt_ad_baselines} to establish reference performance levels on \kaputtrare.
These baselines span supervised classifiers as well as unsupervised anomaly detection methods, illustrating the difficulty of the task under realistic low-prevalence conditions.
\href{https://huggingface.co/datasets/amazon/kaputt/tree/main/kaputt-rare}{\kaputtrare}\ is publicly available as a partition file and will serve as the basis for future VAND challenges with focus on anomaly detection in retail logistics.

\begin{table}[ht]
\centering
\caption{Baseline results on \kaputtrare. \texttt{ViT-S} is fine-tuned on the
\kaputtrare\ training set (normal-heavy, ${\sim}1\%$ defect rate); all other
methods are training-free (zero-/few-shot). \texttt{ViT-S} operates on the full
tray image, whereas \texttt{CLIP}, \texttt{WinCLIP}, and \texttt{PatchCore50}
operate on item crops. Evaluated on the test split (10067 samples;
2089 minor, 1117 major defects). All values in~\%.}
\label{tab:kaputt_ad_baselines}
\resizebox{\columnwidth}{!}{
\begin{tabular}{l c c c}
\toprule
Method & \apany & \apmajor & AUROC \\
\midrule
\texttt{ViT-S}~\cite{oquab2023dinov2} & \textbf{58.38} & \textbf{50.09} & \textbf{72.99} \\
\midrule
\texttt{CLIP}~\cite{Radford_2021_CLIP}                  & 36.20 & 17.15 & 56.05 \\
\texttt{WinCLIP-zero}~\cite{jeong2023_winclip} & 33.87 & 19.11 & 52.30 \\
\texttt{WinCLIP-few}~\cite{jeong2023_winclip}  & 34.05 & 19.29 & 52.41 \\
\texttt{PatchCore50}~\cite{patchcore}      & 35.86 & 17.80 & 54.69 \\
\bottomrule
\end{tabular}
}
\end{table}

\clearpage

{
    \small
    \bibliographystyle{ieeenat_fullname}
    \bibliography{main}
}

\clearpage

\appendix
\section*{Supplemental Material}

\section{Efficiency Metric}
\label{app:efficiency_metric}

Evaluating image anomaly detection models typically involves reporting detection quality, such as F1 score, independently of computational cost.
In practice, however, deployment constraints such as throughput, memory footprint, and power consumption are equally decisive. 
Following, we define a single scalar efficiency score that combines performance, e.g., F1 score, with three normalized efficiency metrics: throughput, memory usage, and power consumption.

\paragraph{Design Goals:}

The proposed efficiency metric is designed to satisfy the following requirements:

\begin{enumerate}
    \item \textbf{Interpretability.} The score must lie in the interval $[0,1]$, with 1 indicating perfect efficiency.
    
    \item \textbf{Balanced sensitivity.} No single component should dominate the score. An improvement in any dimension should be reflected in the final value, while a severe weakness in any dimension should penalize the model.
    
    \item \textbf{Diminishing returns for throughput.} Beyond a service-level target, such as $60$ frames per second, additional speed yields diminishing value. Unlike memory and power, theoretical throughput is unbounded, so a saturating normalization is required.
    
    \item \textbf{Monotonicity.} The score should be monotonically increasing in performance and throughput, and monotonically decreasing in memory usage and power consumption.
\end{enumerate}

\subsection{Metric Definition}

Each raw measurement is mapped to a normalized value in $[0,1]$ so that all four components are commensurable.

\paragraph{Performance:}

Most performance scores, e.g., \segfscoreaverage\ or \apany\ (see main paper), already lie in $[0,1]$, so no further transformation is required for the \vandchallenge:
\begin{equation}
    q_\text{industrial} = \segfscoreaverage, \qquad q_\text{industrial} \in [0,1].
    \label{eq:app_f1}
\end{equation}
\begin{equation}
    q_\text{retail} = \apany, \qquad q_\text{retail} \in [0,1].
\end{equation}

\paragraph{Throughput:}

Let $t$ denote the raw throughput in frames per second.
Throughput is normalized using a diminishing-returns function against a target frame rate $t_{\max}$:
\begin{equation}
    \tau = \frac{t}{t + t_{\max}}, \qquad t \geq 0.
    \label{eq:app_throughput}
\end{equation}

Here, $t_{\max}$ is the service-level target, for example $60$ fps. This mapping is monotonically increasing and bounded. It naturally soft-caps throughput so that doubling an already-fast model contributes less than doubling a slow one.

\paragraph{Memory Usage:}

Let $m$ denote peak GPU memory usage in MB. Memory is linearly normalized against a device budget $m_{\max}$:
\begin{equation}
    \mu = \frac{m}{m_{\max}}, \qquad \mu \in [0,1].
    \label{eq:app_memory}
\end{equation}

Because lower memory usage is preferable, the complement $1-\mu$ is used in the composite score.
Exceeding the budget naturally results in out-of-memory errors and, thus, $m$ cannot be measured.

\paragraph{Power Consumption:}

Let $p$ denote average power draw in watts. Power consumption is normalized against a power budget $p_{\max}$:
\begin{equation}
    \pi = \frac{p}{p_{\max}}, \qquad \pi \in [0,1].
    \label{eq:app_power}
\end{equation}

As with memory, lower power consumption is preferable, so the complement $1-\pi$ is used in the composite score.

\paragraph{Composite Efficiency Score:}

The four normalized components are combined using a geometric mean to obtain the efficiency score $\mathcal{E}$:
\begin{equation}
    \mathcal{E}
    =
    \left[
    q \cdot \tau \cdot (1-\mu) \cdot (1-\pi)
    \right]^{1/4}.
    \label{eq:app_efficiency_score}
\end{equation}

Substituting the component definitions gives:
\begin{equation}
    \mathcal{E}
    =
    \left[
    q
    \cdot
    \frac{t}{t+t_{\max}}
    \cdot
    \left(1-\frac{m}{m_{\max}}\right)
    \cdot
    \left(1-\frac{p}{p_{\max}}\right)
    \right]^{1/4}.
    \label{eq:app_efficiency_score_expanded}
\end{equation}

Based on this formulation, the ideal score for a model that meets the throughput service-level agreement is approximately:
\begin{equation}
    \mathcal{E}_{\mathrm{ideal}} \approx 0.841.
\end{equation}

\paragraph{Rationale for the Geometric Mean:}

The geometric mean is chosen over the arithmetic mean for three reasons.

First, it enforces \textbf{zero-annihilation}. If any single component is zero, such as when the model has $\segfscoreaverage=0$, consumes the entire memory budget, or consumes the entire power budget, then the composite score collapses to zero. This correctly reflects the fact that a completely failing dimension can render the model unusable.

Second, it encourages \textbf{balanced trade-offs}. The geometric mean penalizes imbalance. For example, a model with component values
\begin{equation}
    (0.9, 0.9, 0.9, 0.9)
\end{equation}
scores higher than a model with values
\begin{equation}
    (1.0, 1.0, 1.0, 0.6),
\end{equation}
(0.9 vs. $\approx0.88$) even though both have the same arithmetic mean (0.9).
This encourages well-rounded models rather than models that excel in only a subset of dimensions.

Third, it preserves a \textbf{bounded output}. Because every factor lies in $[0,1]$, the geometric mean also lies in $[0,1]$, preserving interpretability without requiring an additional sigmoid or min--max rescaling step.

\subsection{Reference Operating Point}

The reference maxima used in the experiments are shown in Table~\ref{tab:app_reference_operating_point}. These values are representative of a mid-range edge deployment scenario.

\begin{table}[h]
    \centering
    \caption{Reference operating-point parameters.}
    \label{tab:app_reference_operating_point}
    \begin{tabular}{lll}
        \hline
        \textbf{Parameter} & \textbf{Value} & \textbf{Unit} \\
        \hline
        $t_{\max}$, target throughput & $30$ & fps \\
        $m_{\max}$, memory budget & $62470.692864$ & MB \\
        $p_{\max}$, power budget & $56.21$ & W \\
        \hline
    \end{tabular}
\end{table}

\subsection{Experimental Analysis}

The behavior of the efficiency score is analyzed through controlled sweeps in which each component varies independently while the remaining three are held at their ideal values:
\begin{equation}
    q=1, \qquad \tau=1, \qquad \mu=0, \qquad \pi=0.
\end{equation}

\subsubsection{Single-Component Sensitivity}

\begin{figure}[ht]
    \centering
    \includegraphics[width=0.9\linewidth]{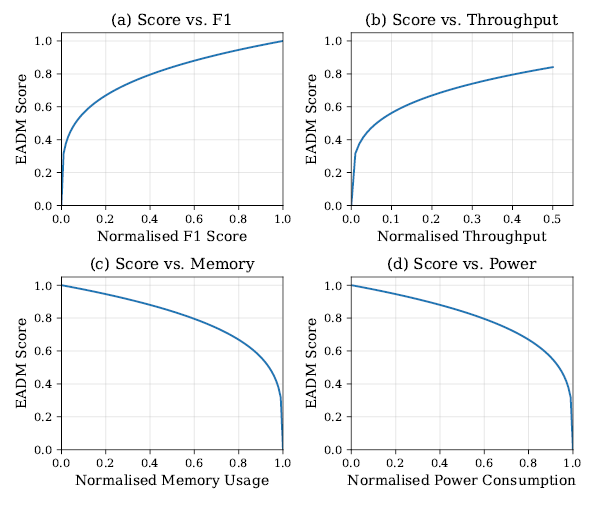}
    \caption{
    Efficiency score sensitivity to each component in isolation.
    (a) F1 score: fourth-root growth, where even modest F1 improvements noticeably raise the score.
    (b) Throughput: diminishing-returns normalization soft-caps the curve well below $1$.
    (c) Memory usage: because the complement $1-\mu$ is used, the score decreases as memory usage grows.
    (d) Power consumption: because the complement $1-\pi$ is used, the score decreases as power consumption grows.
    }
    \label{fig:app_single_component_sensitivity}
\end{figure}

Figure~\ref{fig:app_single_component_sensitivity} shows the efficiency score as each of the four components is swept from its worst to its best value.

For the performance $q$, the score exhibits fourth-root growth:
\begin{equation}
    \mathrm{EfficiencyScore} \propto q^{1/4}.
\end{equation}

Because $q^{1/4}$ has a steep initial gradient, even small improvements in performance at the low end produce meaningful score gains.
This correctly reflects that a model with $q=0.3$ is far more useful than one with $q=0.1$.

For throughput, the score saturates due to the diminishing-returns normalization in Equation~\ref{eq:app_throughput}.
The maximum possible throughput contribution for a selected service level agreement is:
\begin{equation}
    0.5^{1/4} \approx 0.84,
\end{equation}
which enforces the design intent that raw speed alone cannot dominate the score.

For memory and power, the score decreases monotonically as normalized resource usage increases, reaching zero when the full budget is consumed.

\subsubsection{Composite Sweep}

\begin{figure}[h]
    \centering
    \includegraphics[width=0.75\linewidth]{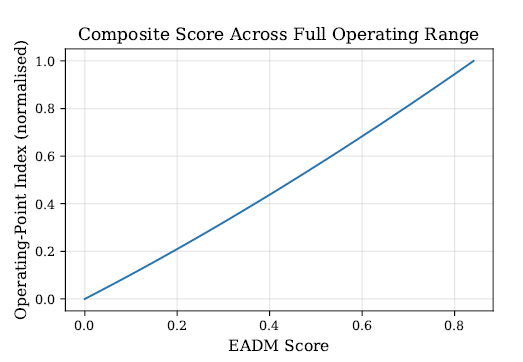}
    \caption{
    Composite efficiency score across the full operating range.
    The approximately linear response confirms that no single component dominates the metric.
    }
    \label{fig:app_composite_sweep}
\end{figure}

Figure~\ref{fig:app_composite_sweep} shows the efficiency score when all four components are swept simultaneously from their worst to their best values. In this sweep, F1 and throughput increase linearly while memory and power decrease linearly.

The resulting curve is approximately linear, confirming that the metric distributes sensitivity evenly across the four dimensions.

\subsubsection{Joint F1--Throughput Surface}

\begin{figure}[h]
    \centering
    \includegraphics[width=0.75\linewidth]{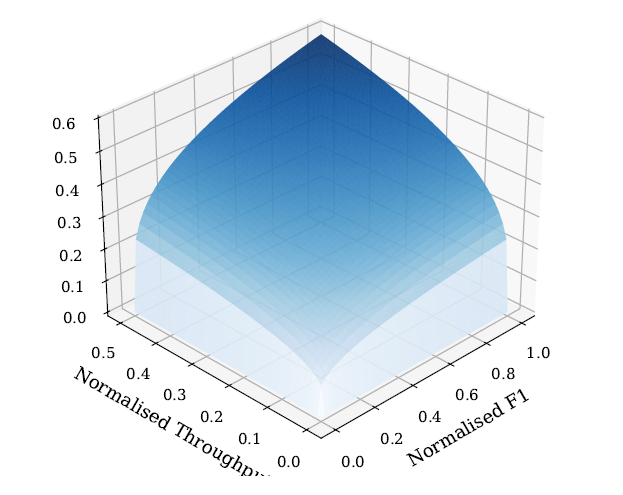}
    \caption{
    Efficiency score surface over F1 and throughput at moderate resource usage, with memory $=50\%$ and power $=50\%$.
    The concave surface illustrates the balanced trade-off between detection quality and inference speed.
    }
    \label{fig:app_f1_throughput_surface}
\end{figure}

Figure~\ref{fig:app_f1_throughput_surface} shows the score surface as F1 score and throughput vary jointly, with memory and power fixed at $50\%$ of their respective budgets:
\begin{equation}
    \mu = 0.5, \qquad \pi = 0.5.
\end{equation}

The surface shows that achieving a high composite score requires both good detection quality and reasonable throughput. Excelling in only one axis is insufficient.

\subsection{Discussion}

\subsubsection{Configurable Reference Points}

The parameters $t_{\max}$, $m_{\max}$, and $p_{\max}$ are intentionally left as user-configurable constants. Different deployment targets, such as cloud GPUs, edge CPUs, or FPGAs, can set these values to match their constraints without changing the functional form of the metric.

\subsubsection{Extensibility}

Additional efficiency dimensions, such as model size in parameters or latency jitter, can be incorporated by extending the geometric mean to $n$ factors and adjusting the exponent to $1/n$.

For normalized components $c_1, c_2, \ldots, c_n$, the generalized score is:
\begin{equation}
    \mathrm{EfficiencyScore}
    =
    \left(
    \prod_{i=1}^{n} c_i
    \right)^{1/n}.
    \label{eq:app_generalized_efficiency_score}
\end{equation}

\subsection{Limitations}

The proposed metric has several limitations.

First, the geometric mean assigns equal weight to all four components. Practitioners who consider detection quality more important than efficiency may wish to introduce a weighted geometric mean:
\begin{equation}
    \mathcal{E}_{w}
    =
    \prod_{i=1}^{4} c_i^{w_i},
    \qquad
    \sum_{i=1}^{4} w_i = 1,
    \label{eq:app_weighted_efficiency_score}
\end{equation}
where $c_i$ are the normalized components and $w_i$ are user-defined weights.

Second, the throughput normalization saturates after $0.5$.
For applications where extreme speed is critical, an alternative sublinear normalization, such as a logarithmic function, may be preferable.

\end{document}